\documentclass[11pt]{article}

\usepackage[]{ACL2023}

\usepackage{times}
\usepackage{latexsym}

\usepackage[T1]{fontenc}

\usepackage[utf8]{inputenc}

\usepackage{microtype}

\usepackage{color}
\definecolor{my-color}{rgb}{0.28, 0.58, 0.28}

\usepackage{graphicx}
\usepackage{booktabs}
\usepackage{multirow}
\usepackage{adjustbox}
\usepackage{amssymb}
\usepackage{colortbl}
\usepackage{diagbox}
\usepackage{float}
\usepackage{graphicx}
\usepackage{amsmath}
\usepackage{algorithm} 
\usepackage{algpseudocode}

\definecolor{red1}{RGB}{253, 235, 232}
\definecolor{red2}{RGB}{243, 173, 172}
\definecolor{red3}{RGB}{228, 121, 121}

\definecolor{green1}{RGB}{209, 226, 255}
\definecolor{green2}{RGB}{163, 197, 255}
\definecolor{green3}{RGB}{121, 170, 255}

\definecolor{Blue}{HTML}{2020df}
\definecolor{Pink}{HTML}{F08080}

\usepackage{inconsolata}

\title{Neuron Specialization: Leveraging Intrinsic \\ Task Modularity for Multilingual Machine Translation}

\author{Shaomu Tan \qquad Di Wu \qquad Christof Monz\\
  Language Technology Lab\\ University of Amsterdam \\
  \texttt{\{s.tan, d.wu, c.monz\}@uva.nl} \\}

\begin{document}
\maketitle
\begin{abstract}

Training a unified multilingual model promotes knowledge transfer but inevitably introduces \textit{negative interference}. Language-specific modeling methods show promise in reducing interference. However, they often rely on heuristics to distribute capacity and struggle to foster cross-lingual transfer via isolated modules. In this paper, we explore intrinsic task modularity within multilingual networks and leverage these observations to circumvent interference under multilingual translation. We show that neurons in the feed-forward layers tend to be activated in a language-specific manner. Meanwhile, these specialized neurons exhibit structural overlaps that reflect language proximity, which progress across layers. Based on these findings, we propose \textit{Neuron Specialization}, an approach that identifies specialized neurons to modularize feed-forward layers and then continuously updates them through sparse networks. Extensive experiments show that our approach achieves consistent performance gains over strong baselines with additional analyses demonstrating reduced interference and increased knowledge transfer.\footnote{We release code at \url{https://anonymous.4open.science/r/NS-3D93}}



\end{abstract}

\section{Introduction}


Jointly training multilingual data in a unified model with a shared architecture for different languages has been a trend ~\citep{conneau2020unsupervised, le2022bloom} encouraging knowledge transfer across languages, especially for low-resource languages~\cite{johnson2017google, pires2019multilingual}. 
However, such a training paradigm also leads to \textit{negative interference} due to conflicting optimization demands~\cite{wang2020gradient}. This interference often causes performance degradation for high-resource languages~\cite{li2021robust, pfeiffer2022lifting} and can be further exacerbated by limited model capacity~\cite{shaham-etal-2023-causes}.


Modular-based methods, such as Language-specific modeling~\cite{zhang2020improving} and adapters~\cite{bapna2019simple}, aim to mitigate interference by balancing full parameter sharing with isolated or partially shared modules~\cite{pfeiffer2023modular}. 
However, they heavily depend on heuristics for allocating task-specific capacity and face challenges in enabling knowledge transfer between modules~\cite{zhang2020share}. 
Specifically, such methods rely on prior knowledge for managing parameter sharing such as language-family adapters~\cite{chronopoulou2023language} or directly isolate parameters per language, which impedes transfer~\cite{pires2023learning}.


Research in vision and cognitive science has shown that unified multi-task models may spontaneously develop task-specific functional specializations for distinct tasks~\cite{yang2019task, dobs2022brain}, a phenomenon also observed in mixture of experts Transformer systems~\cite{zhang-etal-2023-emergent}. 
These findings suggest that through multi-task training, networks naturally evolve towards specialized modularity to effectively manage diverse tasks, with the ablation of these specialized modules adversely affecting task performance~\cite{pfeiffer2023modular}. 
Despite these insights, exploiting the inherent structural signals for multi-task optimization remains largely unexplored.

In this work, we explore the intrinsic task-specific modularity within multi-task networks in Multilingual Machine Translation (MMT), treating each language pair as a separate task. 
We focus on analyzing the intermediate activations in the Feed-Forward Networks (FFN) where most model parameters reside. 
Our analysis shows that neurons activate in a language-specific way, yet they present structural overlaps that indicate language proximity. 
Moreover, this pattern evolves across layers in the model, consistent with the transition of multilingual representations from language-specific to language-agnostic~\cite{kudugunta2019investigating}.

Building on these observations, we introduce \textit{Neuron Specialization}, a novel method that leverages intrinsic task modularity to reduce interference and enhance knowledge transfer. 
In general, our approach selectively updates the FFN parameters during back-propagation for different tasks to enhance task specificity. 
Specifically, we first identify task-specific neurons from pre-trained multilingual translation models, using standard forward-pass validation processes without decoding. 
We then specifically modularize FFN layers using these specialized neurons and continuously update FFNs via sparse networks.


Extensive experiments on small- (IWSLT) and large-scale EC30~\cite{tan2023towards} multilingual translation datasets show that our method consistently achieves performance gains over strong baselines. 
Moreover, we conduct in-depth analyses to demonstrate that our method effectively mitigates interference and enhances knowledge transfer in high and low-resource languages, respectively. 
Our main contributions are summarized as follows:

\begin{itemize}

  \item We identify inherent multilingual modularity by showing that neurons activate in a language-specific manner and their overlapping patterns reflect language proximity.
  

  \item Building on these findings, we enhance task specificity through sparse sub-networks, achieving consistent improvements in translation quality over strong baselines.
  

  \item We employ analyses to show that our method effectively reduces interference in high-resource languages and boosts knowledge transfer in low-resource languages.



\end{itemize}


\section{Related Work}

\paragraph{Multilingual Interference.}

Multilingual training enables knowledge transfer but also causes \textit{interference}, largely due to optimization conflicts among various languages or tasks~\cite{wang2022parameter}. Methods addressing conflicts between tasks hold promise to reduce interference~\cite{wang2020gradient}, yet they show limited effectiveness in practical applications~\cite{xin2022current}. Scaling up model size reduces interference directly but may lead to overly large models~\cite{chang2023multilinguality}, with risks of overfitting~\cite{aharoni2019massively}.

\paragraph{Language-Specific Modeling.}


Modular-based approaches enhance the unified model by adding language-dependent modules such as adapters~\cite{bapna2019simple} or language-aware layers~\cite{zhang2020improving}. Although the unified model serves as a common foundation, these approaches struggle to facilitate knowledge transfer among isolated modules due to a lack of clear inductive biases and thus heavy reliance on heuristics. For instance, ~\citet{chronopoulou2023language} rely on priori knowledge to control parameter sharing in language family adapters,~\citet{bapna2019simple,pires2023learning} isolate modules per language, hindering knowledge sharing.



Additionally, these modular-based methods substantially increase the number of parameters, thereby leading to increased memory demands and slower inference times~\citep{liao-etal-2023-parameter, liao2023make}. 
Despite adapters normally being lightweight, they can easily accumulate to a significant parameter growth when dealing with many languages. 
In contrast, our method leverages the model's intrinsic modularity signals to promote task separation, without adding extra parameters.

\paragraph{Sub-networks in Multi-task Models.} The lottery ticket hypothesis~\cite{frankle2018lottery} states that within dense neural networks, sparse subnetworks can be found with iterative pruning to achieve the original network's performance. 
Following this premise, recent studies attempt to isolate sub-networks of a pre-trained unified model that captures task-specific features~\cite{lin2021learning,he2023gradient,choenni2023cross}. 
Nonetheless, unlike our method that identifies intrinsic modularity within the model, these approaches depend on fine-tuning to extract the task-specific sub-networks. 
This process may not reflect the original model modularity and also can be particularly resource-consuming for multiple tasks.

Specifically, these methods extract the task-specific sub-networks by fine-tuning the original unified multi-task model on specific tasks, followed by employing pruning to retain only the most changed parameters. We argue that this process faces several issues: 1) The sub-network might be an artifact of fine-tuning, suggesting the original model may not inherently possess such modularity. 2) This is further supported by the observation that different random seeds during fine-tuning lead to varied sub-networks and performance instability~\cite{choenni2023cross}. 3) The process is highly inefficient for models covering multiple tasks, as it necessitates separate fine-tuning for each task.

\section{Neuron Structural Analysis}\label{sec:Neuron_Structural_Analyasis}


Recent work aims to identify a subset of parameters within pre-trained multi-task networks that are sensitive to distinct tasks. 
This exploration is done by either 1) ablating model components to assess impacts on performance, such as~\citet{dobs2022brain} ablate task-specific filters in vision models by setting their output to zero; or 2) fine-tuning the unified model on task-specific data to extract sub-networks~\cite{lin2021learning,he2023gradient,choenni2023examining}. 
These approaches, however, raise a fundamental question, namely whether the modularity is inherent to the original model, or simply an artifact introduced by network modifications.

In this paper, we perform a thorough identification of task-specific modularity through the lens of neuron behaviors, without altering the original parameters or architectures. We focus on the neurons --- the intermediate activations inside the Feed-Forward Networks (FFN) --- to investigate if they indicate task-specific modularity features. As FFN neurons are active (>0) or inactive (=0) due to the $\mathit{ReLU}$ activation function, this binary activation state offers a clear view of their contributions to the network's output. 
Intuitively, neurons that remain inactive for one task but show significant activation for another may be indicative of specialization for the latter. 
Analyzing such modularity structures can improve our understanding of fundamental properties in multi-task models and yield insights to advance multi-task learning.

\subsection{Identifying Specialized Neurons}\label{sec:Neuron_Analysis:Identification}

We choose multilingual translation as a testbed, treating each translation direction as a distinct task throughout the paper. 
We start with a pre-trained multilingual model with $d_{\mathit{ff}}$ as its dimension of the FFN layer.
We hypothesize the existence of neuron subsets specialized for each task and describe the identification process of an FFN layer as follows.

\paragraph{Activation Recording.}
Given a validation dataset $D_t$ for the $t$-th task, we measure activation frequencies in an FFN layer during validation.
For each sample $x_i \in D_t$, we record the state of each neuron after $\mathit{ReLU}$, reflecting whether the neuron is active or inactive to the sample. 
We use a binary vector $a^{t}_{i} \in \mathbb{R}^{d_{\mathit{ff}}}$ to store this neuron state information. 
Note that this vector aggregates neuron activations for all tokens in the sample by taking the neuron union of them.
By further merging all of the binary vectors for all samples in $D_t$, an accumulated vector $a^{t} = \sum_{x_i \in D_{t}} a^{t}_{i} $ can be derived, which denotes the frequency of each neuron being activated during a forward pass given a task-specific dataset $D_t$.


\paragraph{Neuron Selection.} We identify specialized neurons for each task $t$ based on their activation frequency $a^{t}$. 
A subset of neurons $S_{k}^{t}$ is progressively selected based on the highest $a^{t}$ values until reaching a predefined threshold $k$, where

\begin{equation}\label{equation:threshold}
    \sum_{i \in S_{k}^{t}} a^{t}_{(i)} >= k \sum_{i=1}^{d_{\mathit{ff}}} a^{t}_{(i)}
\end{equation}

Here, the value $a_{(i)}^{t}$ is the frequency of the activation at dimension $i$, and $\sum_{i=1}^{d_{\mathit{ff}}} a^{t}_{(i)}$ is the total activation of all neurons for an FFN layer. 
$k$ is a threshold factor, varying from 0\% to 100\%, indicating the extent of neuron activation deemed necessary for specialization. 
A lower $k$ value results in higher sparsity in specialized neurons; $k=0$ means no neuron will be involved, while $k=100$ fully engages all neurons, the same as utilizing the full capacity of the original model. 
This dynamic approach emphasizes the collective significance of neuron activations up to a factor of $k$. 
In the end, we repeat these processes to obtain the specialized neurons of all FFN layers for each task.

\begin{figure*}[h]
    \centering
    \includegraphics[width=0.95\linewidth]{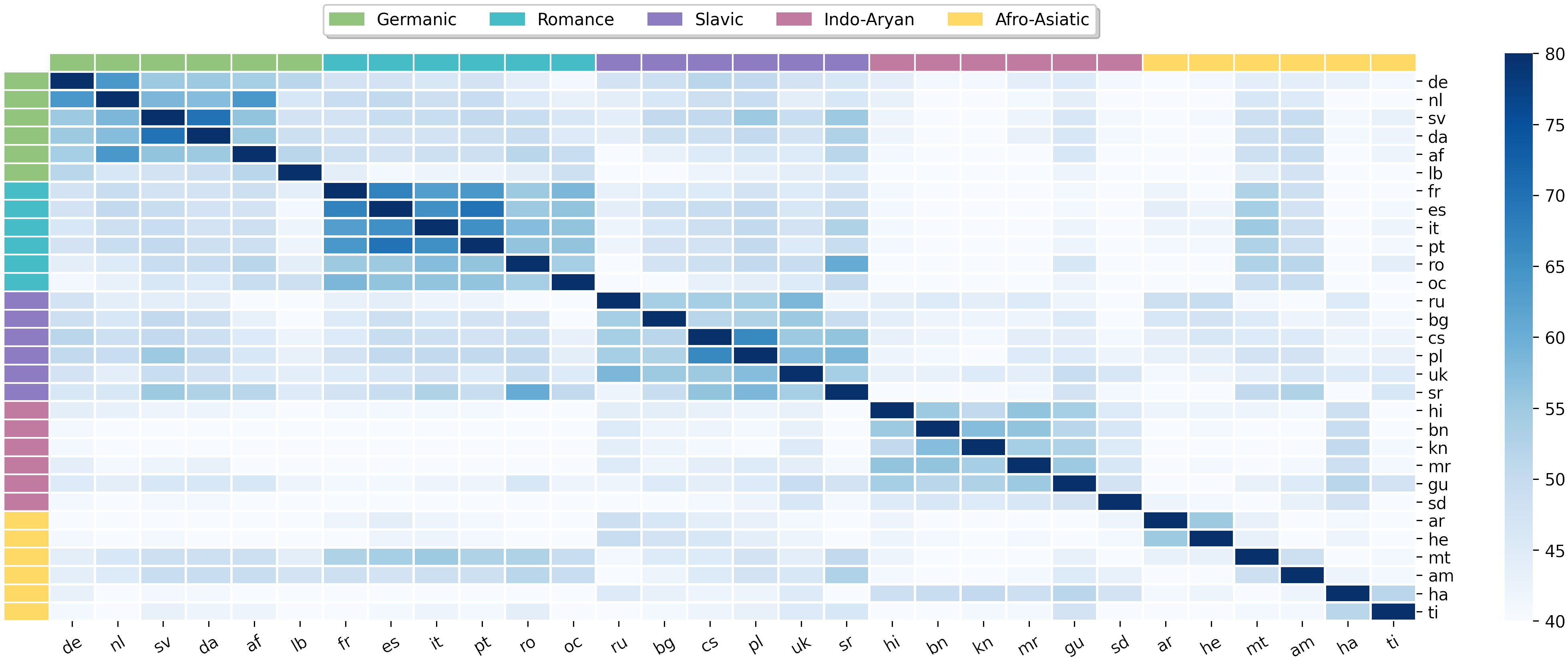}
    \caption{Pairwise Intersection over Union (IoU) scores for specialized neurons extracted from the first decoder FFN layer across all out-of-English translation directions to measure the degree of overlap. Darker cells indicate stronger overlaps, with the color threshold set from 40 to 80 to improve visibility.}
    \label{fig:overlap_teaser}
\end{figure*}

\subsection{Analysis on EC30}

In this section, we describe how we identify specialized neurons on EC30~\cite{tan2023towards}, where we train an MMT model covering all directions. 
EC30 is a multilingual translation benchmark that is carefully designed to consider diverse linguistic properties and real-world data distributions. 
It collects high to low-resource languages, resulting in 30 diverse languages from 5 language families, allowing us to connect our observations with linguistic properties easily. 
See Sections~\ref{sec:Experimental_Setup} for details on data and models.

\subsubsection{Neuron Overlaps Reflect Language Proximity}

We identified specialized neurons following Section~\ref{sec:Neuron_Analysis:Identification}, while setting the cumulative activation threshold $k$ at 95\%. 
This implies that the set of specialized neurons covers approximately 95\% of the total activations. 
Intuitively, two similar tasks should have a high overlap between their specialized neuron sets. 
Therefore, we examined the overlaps among specialized neurons across different tasks by calculating the Intersection over Union (IoU) scores: For task $t_{i}$ and $t_{j}$, with specialized neurons denoted as sets $S^i$ and $S^j$, their overlap is quantified by $\text{IoU}(S^i, S^j) = \frac{|S^{i} \cap S^{j}|}{|S^{i} \cup S^{j}|} $.



Figure~\ref{fig:overlap_teaser} shows the IoU scores for specialized neurons across different tasks in the first decoder layer.
Figures for the other layers can be found in Appendix~\ref{appendix:Visualizations}. 
We first note a structural separation of neuron overlaps, indicating a preference for language specificity. 
Notably, neuron overlap across language families is relatively low, a trend more pronounced in encoder layers (Figure~\ref{fig:iou_overall_enc_first}). 
Secondly, this structural distinction generally correlates with language proximity as indicated by the clustering pattern in Figure~\ref{fig:overlap_teaser}. 
This implies that target languages from the same family are more likely to activate similar neurons in the decoder, even when they use different writing systems, e.g., Arabic (ar) and Hebrew (he).
Overlaps also show linguistic traits beyond family ties, exemplified by notable overlaps between Maltese (mt) and languages in the Romance family due to vocabulary borrowing.

\begin{figure}[h!]
    \centering
    \includegraphics[width=\linewidth]{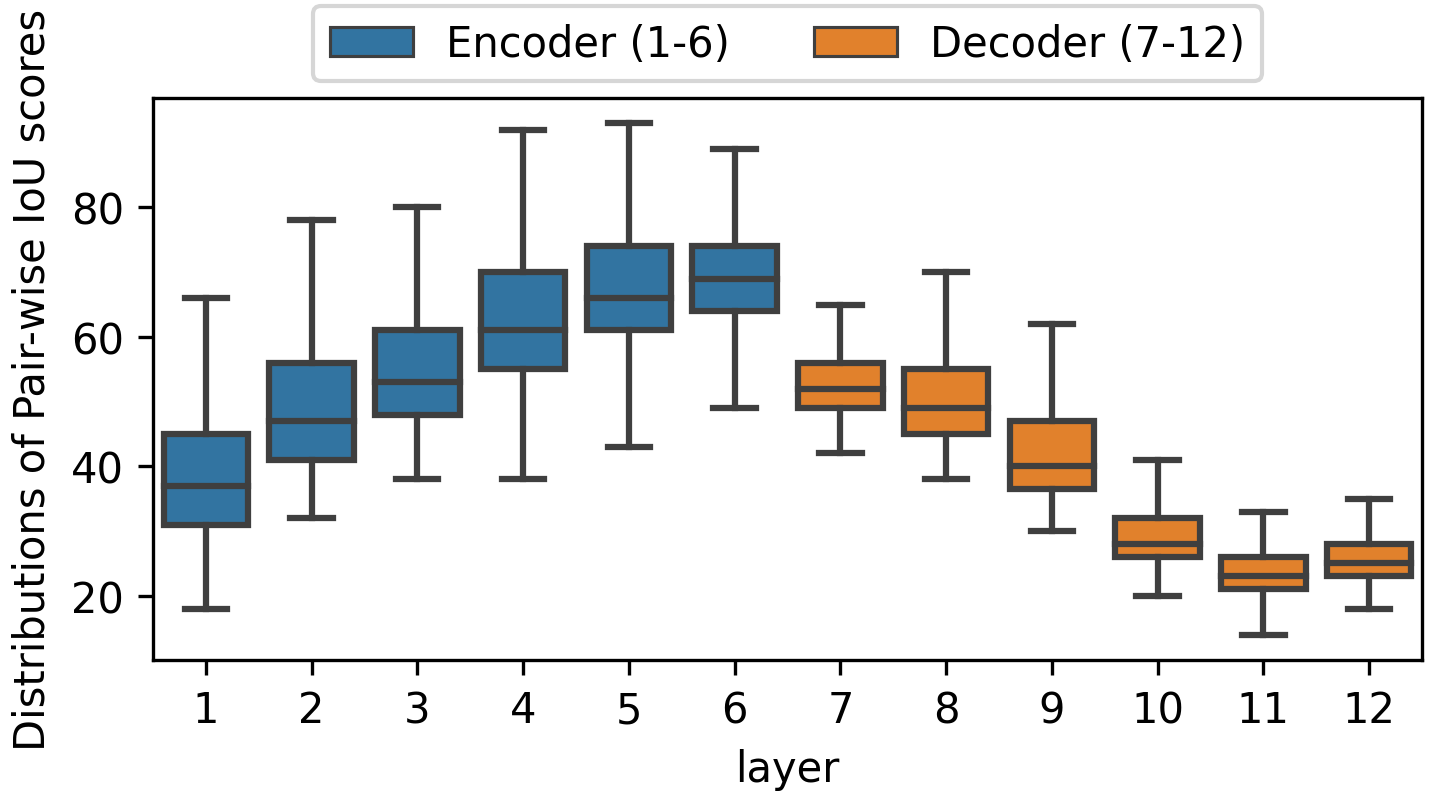}
    \caption{Progression of distribution of IoU scores for specialized neurons across layers on the EC30 dataset.
    The scores are measured for different source and target languages in the Encoder and Decoder, respectively.
    }
    \label{fig:overlap_layers}
\end{figure}

\subsubsection{The Progression of Neuron Overlaps}\label{sec:Neuron_Structural_Analyasis:neuron_overlap_progression}

To analyze how specialized neuron overlaps across tasks evolve within the model, we visualize the IoU score distribution across layers in Figure~\ref{fig:overlap_layers}. 
For each layer, we compute the pair-wise IoU scores between all possible tasks and then show them in a distribution.
Overall, we observe that from shallow to deeper layers, structural distinctions intensify in the decoder (decreasing IoU scores) and weaken in the encoder (increasing IoU scores).



On the one hand, all neuron overlaps increase as we move up the encoder, regardless of whether these tasks are similar or not.
This observation may suggest that the neurons in the encoder become more language-agnostic, as they attempt to map different scripts into semantic concepts. 
As for the Decoder, the model presents intensified modularity in terms of overlaps of specialized neurons. 
This can be seen by all overlaps becoming much smaller, indicating that the neurons behave more separately. 
Additionally, we found the progression of neuron overlaps is similar to the evolution of multilingual representation: embedding gets closer in the encoder and becomes more dissimilar in the decoder~\cite{kudugunta2019investigating}. 
Our observations, highlighting the inherent features of the multilingual translation model, occur without modifying the network's outputs or parameters.

\section{Neuron Specialization Training}

Our neuron structural analysis showed the presence of specialized neurons within the Feed-Forward Network (FFN) layers of a multilingual network. 
We hypothesize that continuously training the model, while leveraging these specialized neurons' intrinsic modular features, can further enhance task-specific performance. 
Building on this hypothesis, we propose \textit{Neuron Specialization}, an approach that leverages specialized neurons to modularize the FFN layers in a task-specific manner. 


\subsection{Vanilla Feed-Forward Network}

We first revisit the Feed-Forward Network (FFN) in Transformer~\cite{vaswani2017attention}. 
The FFN, crucial to our analysis, consists of two linear layers (fc1 and fc2) with a $\mathit{ReLU}$ activation function. 
Specifically, the FFN block first processes the hidden state $H \in \mathbb{R}^{n\times d}$ ($n$ denotes number of tokens in a batch) through fc1 layer $W_{1} \in \mathbb{R}^{d\times d_{\mathit{ff}}}$. 
Then the output is passed to $\mathit{ReLU}$ and the fc2 layer $W_{2}$, as formalized in Eq~\ref{equation:FFN_transformer}, with bias terms omitted.


\begin{equation}\label{equation:FFN_transformer}
\mathrm{FFN}(H) = \mathrm{ReLU}(HW_{1})\,W_{2}.
\end{equation}

\subsection{Specializing Task-Specific FFN}

Next, we investigate continuous training upon a subset of specialized parameters within FFN for each task. 
Given a pre-trained vanilla multilingual Transformer model with tags to identify the language pairs, e.g.,~\citet{johnson2017google}, we can derive specialized neuron set $S_{k}^{t}$ for each layer of a task task\footnote{We treat each translation direction as a distinct task.} $t$ and threshold $k$ following the method outlined in Section~\ref{sec:Neuron_Analysis:Identification}. 
Then, we derive a boolean mask vector \mbox{$m_{k}^{t} \in \{0, 1\}^{d_{\mathit{ff}}}$} from $S_{k}^{t}$, where the $i$-th element in $m_{k}^{t}$ is set to 1 only when $i \in S_{k}^{t}$, and apply it to control parameter updates. 
Specifically, we broadcast $m_{k}^{t}$ and perform Hadamard Product with $W_{1}$ in each FFN layer as follows:

\begin{equation}\label{equation:FFN_SN_transformer}
\textit{FFN}(H) = \mathit{ReLU}(H(m_{k}^{t} \odot W_{1})) W_{2}.
\end{equation}

$m_{k}^{t}$ plays the role of controlling parameter update, where the boolean value of $i$-th element in $m_{k}^{t}$ denotes if the $i$-th row of parameters in $W_{1}$ can be updated or not for each layer\footnote{Note that $m_{k}^{t}$ is layer-specified, we drop layer indexes hereon for simplicity of notation.} during continues training. 
Broadly speaking, our approach selectively updates the first FFN (fc1) weights during back-propagation, tailoring the model more closely towards specific translation tasks and reinforcing neuron separation. 
Note that while fc1 is selectively updated for specific tasks, other parameters are universally updated to maintain stability, and the same masking is applied to inference to ensure consistency. 
We provide the pseudocode of our method in Appendix~\ref{appendix:Pseudocodes}.

\section{Experimental Setup}\label{sec:Experimental_Setup}
In this section, we evaluate the capability of our proposed method on small (IWSLT) and large-scale (EC30) multilingual machine translation tasks. 
More details of the datasets are in Appendix~\ref{appendix:Dataset_Details}.

\subsection{Datasets}
    \paragraph{IWSLT.} Following~\citet{lin2021learning}, we constructed an English-centric dataset with eight languages using IWSLT-14, ranging from 89k to 169k in corpus size. 
    We learned a 30k SentencePiece unigram~\cite{kudo2018sentencepiece} shared vocabulary and applied temperature oversampling with $\tau = 2$ to balance low-resource languages. 
    For a more comprehensive evaluation, we replaced the standard test set with Flores-200~\cite{costa2022no}, merging \textit{devtest} and \textit{test}, which offers multiple parallel sentences per source text.

    \paragraph{EC30.}\label{EC30} We further validate our methods using the large-scale EC30 dataset~\cite{tan2023towards}, which features 61 million parallel training sentences across 30 English-centric language pairs, representing five language families and various writing systems. 
    We classify these language pairs into low-resource (=100k), medium-resource (=1M), and high-resource (=5M) categories. Following~\citet{wu-monz-2023-beyond}, we build a 128k size shared SentencePiece BPE vocabulary. 
    Aligning with the original EC30 setups, we use Ntrex-128~\cite{federmann2022ntrex} as the validation set. Also, we use Flores-200 (merging \textit{devtest} and \textit{test}) as test sets for cross-domain evaluation.

\begin{table*}[h!]
\centering
\def\arraystretch{1.00}
\resizebox{0.85\linewidth}{!}{%
\begin{tabular}{c|c|rrrrrrrrr}
\toprule
\multicolumn{1}{c|}{Language}& 
\multirow{2}{*}{$\Delta \theta$} & \multicolumn{1}{c}{Fa} & \multicolumn{1}{c}{Pl} & \multicolumn{1}{c}{Ar} & \multicolumn{1}{c}{He} & \multicolumn{1}{c}{Nl} & \multicolumn{1}{c}{De} & \multicolumn{1}{c}{It} & \multicolumn{1}{c}{Es} &  \multirow{2}{*}{Avg}\\
\multicolumn{1}{c|}{Size}& 
\multicolumn{1}{c|}{} & 
\multicolumn{1}{c}{89k} & 
\multicolumn{1}{c}{128k} & \multicolumn{1}{c}{139k} & \multicolumn{1}{c}{144k} & \multicolumn{1}{c}{153k} & \multicolumn{1}{c}{160k} & \multicolumn{1}{c}{167k} & \multicolumn{1}{c}{169k} & \\
\midrule
\rowcolor[gray]{0.9}
\multicolumn{11}{c}{One-to-Many (O2M / En-X)} \\
mT-small  & - & 14.5 & 9.9  & 12.0 & 13.1 & 17.0 & 20.6 & 17.3 & 18.3 & 15.4 \\
Adapter$_{LP}$ & +67\% &  +0.1 & -0.1 & +0.4 & \textbf{+1.4} & \textbf{+0.2} & +0.6 & +0.1 & \textbf{+0.4} & \textbf{+0.4} \\
LaSS  & 0\% & -2.6 & 0 & +0.6 & +0.7 & -0.2 & \textbf{+0.7} & -0.2 & -0.4 & -0.2 \\
Ours & 0\% & \textbf{+0.7} & \textbf{+0.1} & \textbf{+0.9} & +0.6 & +0.1 & +0.1 & \textbf{+0.2} & -0.3 & +0.3 \\
\midrule
\rowcolor[gray]{0.9}
\multicolumn{11}{c}{Many-to-One (M2O / X-En)} \\
mT-small & - & 19.1 & 19.4 & 25.7 & 30.9 & 30.6  & 28.1 & 29.0 & 34.0 & 24.7 \\
Adapter$_{LP}$ & +67\% & +0.9 & +0.6 & +0.9 & +1.0 & +0.8 & +1.0 & +0.9 & +0.3 & +0.8 \\
LaSS & 0\% & +1.2 & +0.6 & +0.9 & +1.4 & +1.1 & +1.6 & +1.6 & +0.8 & +1.2 \\
Ours & 0\% & \textbf{+1.6} & \textbf{+1.2} & \textbf{+1.7}  & \textbf{+2.0} & \textbf{+1.9} & \textbf{+2.1} & \textbf{+1.8} & \textbf{+1.4} & \textbf{+1.7} \\

\bottomrule 
\end{tabular}%
}
\caption{Average BLEU improvements over the baseline (mT-small) model on the IWSLT dataset. 
$\Delta \theta$ denotes the relative parameter increase over the baseline, encompassing all translation directions. 
The best results are in \textbf{bold}.}
\label{tab:iwslt14}
\end{table*}

\subsection{Systems}

    We compare our method with strong open-source baselines that share similar motivations in reducing interference for multilingual translation tasks.

    \paragraph{Baselines:} 
    \begin{itemize}
    \item \textbf{mT-small.} For IWSLT, we train an mT-small model on Many-to-Many directions as per~\cite{lin2021learning}: a 6-layer Transformer with 4 attention heads, $d$ = 512, $d_{\mathit{ff}}$ = 1,024.

    \item \textbf{mT-big.}\label{mT-big} For EC30, we train a mT-big on Many-to-Many directions following~\citet{wu-monz-2023-beyond}. It has 6 layers, with 16 attention heads, $d$ = 1,024, and $d_{\mathit{ff}}$ = 4,096.
    \end{itemize}

    \paragraph{Adapters.}We employ two adapter methods: 1) Language Pair Adapter (\textbf{Adapter$_{\textit{LP}}$}) and 2) Language Family Adapter (\textbf{Adapter$_{\textit{Fam}}$}). We omit Adapter$_{\textit{Fam}}$ for IWSLT due to its limited languages. Adapter$_{\textit{LP}}$ inserts adapter modules based on language pairs, demonstrating strong effects in reducing interference while presenting no parameter sharing~\cite{bapna2019simple}. In contrast, Adapter$_{\textit{Fam}}$~\cite{chronopoulou2023language} facilitates parameter sharing across similar languages by training modules for each language family. 
    Their bottleneck dimensions are 128 and 512 respectively. 
    See Appendix~\ref{appendix:Model_Training_Details} for more training details.


    \paragraph{LaSS.}~\citet{lin2021learning} proposed LaSS to locate language-specific sub-networks following the lottery ticket hypothesis, i.e., finetuning all translation directions from a pre-trained model and then pruning based on magnitude. 
    They then continually train the pre-trained model by only updating the sub-networks for each direction. 
    We adopt the strongest LaSS configuration by applying sub-networks for both attention and FFNs.
    



\subsection{Implementation and Evaluation}

    We train our baseline models following the same hyper-parameter settings in~\citet{lin2021learning} and~\citet{wu-monz-2023-beyond}. 
    Specifically, we use the Adam optimizer ($\beta1 = 0.9$, $\beta2 = 0.98$, $\epsilon $ = $10^{-9}$) with 5e-4 learning rate and 4k warmup steps in all experiments. 
    We use 4 NVIDIA A6000 (48G) GPUs to conduct most experiments and implement them based on Fairseq~\cite{ott2019fairseq} with FP16. We list detailed training and model specifications for all systems in Appendix~\ref{appendix:Model_Training_Details}.

    We adopt the tokenized BLEU~\cite{papineni2002bleu} for the IWSLT dataset and detokenized case-sensitive SacreBLEU\footnote{nrefs:1|case:mixed|eff:no|tok:13a|smooth:exp|version:2.3.1}~\cite{post2018call} for the EC30 dataset in our main result evaluation section. In addition, we provide ChrF++~\cite{popovic2017chrf++} and COMET~\cite{rei2020comet} in Appendix~\ref{appendix:Results}.

\section{Results and Analyses}

\begin{table*}[h!]
\centering
\def\arraystretch{1.0}%
\resizebox{\linewidth}{!}{%
\begin{tabular}{c|c|rrrrrrrrr|rrr}
\toprule
\multirow{2}{*}{Methods}& 
\multirow{2}{*}{$\Delta \theta$} & 
\multicolumn{3}{c}{High (5M)} & \multicolumn{3}{c}{Med (1M)} & \multicolumn{3}{c}{Low (100K)} & \multicolumn{3}{|c}{All (61M)} \\

\cmidrule(lr){3-5} \cmidrule(lr){6-8} \cmidrule(lr){9-11} \cmidrule(lr){12-14}
      & \multicolumn{1}{c|}{}
      & \multicolumn{1}{c}{O2M} & \multicolumn{1}{c}{M2O} & \multicolumn{1}{c}{Avg} 
      & \multicolumn{1}{c}{O2M} & \multicolumn{1}{c}{M2O} & \multicolumn{1}{c}{Avg} 
      & \multicolumn{1}{c}{O2M} & \multicolumn{1}{c}{M2O} & \multicolumn{1}{c}{Avg} 
      & \multicolumn{1}{|c}{O2M} & \multicolumn{1}{c}{M2O} & \multicolumn{1}{c}{Avg} \\
 \midrule
mT-big & - & 28.1 & 31.6 & 29.9 & 29.7 & 31.6 & 30.6 & 18.9 & 26.0 & 22.4 & 25.5 & 29.7 & 27.7\\


Adapter$_{\textit{Fam}}$ & +70\% &+0.7 & +0.3 & +0.5 & +0.7 & +0.3 & +0.5 & +1.1 & +0.5 & +0.8 & +0.8 & +0.4 & +0.6 \\

Adapter$_{\textit{LP}}$ & +87\% &+1.6 & +0.6 & +1.1 & +1.6 & +0.4 & +1.0 & +0.4 & +0.4 & +0.4 & +1.2 & +0.5 & +0.8 \\

LaSS & 0\% & \textbf{+2.3} & +0.8 & +1.5 & \textbf{+1.7} & +0.2 & +1.0 & -0.1 & -1.8 & -1.0 & +1.3 & -0.3 & +0.5 \\

Random & 0\% & +0.9 & -0.5 & +0.2 & +0.5 & -0.7 & -0.2 & -0.3 & -1.5 & -0.9 & +0.5 & -0.9 & -0.2 \\

\midrule
Ours-Enc & 0\% & +1.2 & +1.1 & +1.1 & +1.0 & +1.0 & +1.0 & +0.7 & +0.8 & +0.8 & +1.0 & +1.0 & +1.0 \\

Ours-Dec & 0\% & +1.2 & +1.1 & +1.1 & +0.9 & +1.1 & +1.0 & +0.7 & +1.1 & +0.9 & +0.9 & +1.1 & +1.0 \\

Ours & 0\% & +1.8 & \textbf{+1.4} & \textbf{+1.6} & +1.4 & \textbf{+1.1} & \textbf{+1.3} & \textbf{+1.4} & \textbf{+0.9} & \textbf{+1.2} & \textbf{+1.5} & \textbf{+1.1} & \textbf{+1.3} \\

\bottomrule 
\end{tabular}%
}
\caption{Average SacreBLEU improvements on the EC30 dataset over the baseline (mT-big), categorized by High, Medium, and Low-resource translation directions. 'Random' denotes continually updating the model with randomly selected task-specific neurons. 'Ours-Enc' and 'Ours-Dec' indicate Neuron Specialization applied solely to the Encoder and Decoder, respectively, while 'Ours' signifies the method applied to both components.}
\label{tab:ec30}
\end{table*}

\subsection{Small-Scale Results on IWSLT} 


We show results on IWSLT in Table~\ref{tab:iwslt14}. 
For Many-to-One (M2O) directions, our method achieves an average +1.7 BLEU gain over the baseline, achieving the best performance among all approaches for all languages. 
The Adapter$_{\textit{LP}}$, with a 67\% increase in parameters over the baseline model, shows weaker improvements (+0.8) than our method. 
As for One-to-Many (O2M) directions, we observed weaker performance improvements for all methods. 
While the gains are modest (averaging +0.3 BLEU), our method demonstrates consistent improvements across various languages in general.

\begin{figure}[t]
    \centering
    \includegraphics[width=\linewidth]{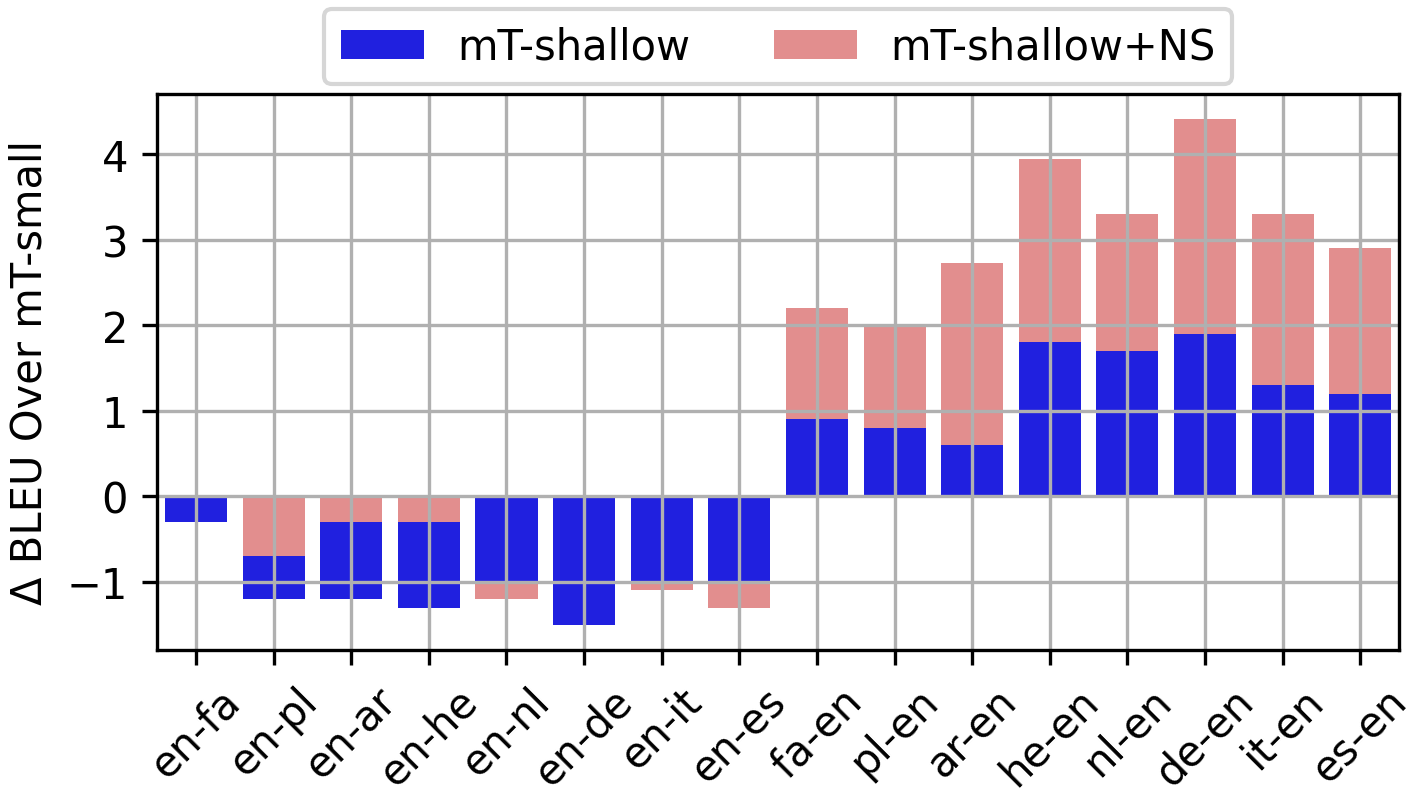}
    \caption{BLEU gains of shallower models over mT-small on IWSLT show improved X-En performance at the expense of En-X. Applying Neuron Specialization reduces EN-X degradation and amplifies X-En gains.}
    \label{fig:shallow}
\end{figure}

\paragraph{Scaling up does not always reduce interference.} ~\citet{shaham-etal-2023-causes,chang2023multilinguality} have found scaling up the model capacity reduces interference, even under low-resource settings. We then investigate the trade-off between performance and model capacity by employing mT-shallow, a shallower version of mT-small with three fewer layers (with $\Delta \theta=-39\%$ for parameters, see Table~\ref{tab:model_spec} for details). Surprisingly, in Figure~\ref{fig:shallow}, we show that reducing parameters improved Many-to-One (X-En) performance but weakened One-to-Many (En-X) results. This result indicates that scaling up the model capacity does not always reduce interference, but may show overfitting to have performance degradation. Furthermore, we show that implementing Neuron Specialization with mT-shallow enhances Many-to-One (X-En) performance in all directions while lessening the decline in One-to-Many (En-X) translation quality in general.

\begin{table*}[!t]
\centering
\resizebox{\linewidth}{!}{
\begin{tabular}{c|rrrrrrrrrr|rr}
\toprule
\multicolumn{1}{c|}{Lang}& 
\multicolumn{1}{r}{De} & 
\multicolumn{1}{r}{Es} & 
\multicolumn{1}{r}{Cs} & 
\multicolumn{1}{r}{Hi} & 
\multicolumn{1}{r}{Ar} & 
\multicolumn{1}{r}{Lb} & 
\multicolumn{1}{r}{Ro} & 
\multicolumn{1}{r}{Sr} & 
\multicolumn{1}{r}{Gu} & 
\multicolumn{1}{r}{Am} & 
\multicolumn{1}{|r}{High} & 
\multicolumn{1}{r}{Low} \\
\multicolumn{1}{c|}{Size} & 
\multicolumn{1}{r}{5m} & 
\multicolumn{1}{r}{5m} & 
\multicolumn{1}{r}{5m} & 
\multicolumn{1}{r}{5m} & 
\multicolumn{1}{r}{5m} & 
\multicolumn{1}{r}{100k} & 
\multicolumn{1}{r}{100k} & 
\multicolumn{1}{r}{100k} & 
\multicolumn{1}{r}{100k} & 
\multicolumn{1}{r}{100k} & 
\multicolumn{1}{|r}{Avg} & 
\multicolumn{1}{r}{Avg} \\
\midrule
\rowcolor[gray]{0.9}
\multicolumn{13}{c}{One-to-Many} \\
Bilingual & 36.3 & 24.6 & 28.7 &43.9 &23.7 &5.5  &16.2 & 17.8 &12.8 &4.1  & 31.8 &11.3 \\ 
mT-big & \colorbox{red3}{-4.7} &\colorbox{red2}{-1.5} & \colorbox{red3}{-3.6} & \colorbox{red3}{-4.4} & \colorbox{red3}{-4.7} & \colorbox{green2}{+9.0} & \colorbox{green2}{+8.9} & \colorbox{green2}{+6.2} & \colorbox{green2}{+13.9} & \colorbox{green2}{+3.1} & \colorbox{red3}{-3.7} & \colorbox{green2}{+8.2} \\
Ours & \colorbox{red2}{-2.0} & \colorbox{red1}{-0.2} & \colorbox{red2}{-1.7} & \colorbox{red2}{-2.4} & \colorbox{red2}{-3.0} & \colorbox{green3}{+10.8} &\colorbox{green3}{+10.0} &\colorbox{green3}{+8.2} &\colorbox{green3}{+16.4} &\colorbox{green3}{+3.7} &\colorbox{red2}{-1.9} & \colorbox{green3}{+9.8}\\
\midrule  
\rowcolor[gray]{0.9}
\multicolumn{13}{c}{Many-to-One} \\
Bilingual & 39.1 & 24.5 & 32.6 & 35.5 & 30.8 & 8.7   & 19.5  & 21.3  & 7.0   & 8.7   & 32.7 & 13.0 \\ 
mT-big & \colorbox{red2}{-1.5} & \colorbox{green1}{+0.9} & \colorbox{green1}{+0.2} & \colorbox{red2}{-1.8} &\colorbox{red2}{-2.3} & \colorbox{green2}{+13.7} &  \colorbox{green2}{+11.9}	& \colorbox{green2}{+10.3}	& \colorbox{green2}{+18.2} & \colorbox{green2}{+12.5} & \colorbox{red2}{-1.1} & \colorbox{green2}{+13.3} \\ 

Ours & \colorbox{red1}{-0.3} & \colorbox{green2}{+1.7} & \colorbox{green2}{+1.8} & \colorbox{red1}{-0.2} & \colorbox{red1}{-0.3} & \colorbox{green3}{+15.3} & \colorbox{green3}{+12.4} & \colorbox{green3}{+11.3} & \colorbox{green3}{+19.6} & \colorbox{green3}{+14.1} & \colorbox{green1}{+0.3} & \colorbox{green3}{+14.5} \\

\bottomrule
\end{tabular}}
\caption{SacreBLEU score comparisons for Multilingual baseline and Neuron Specialization models against Bilingual ones on the EC30 dataset, limited to 5 high- and low-resource languages due to computational constraints.
\colorbox{red1}{Red} signifies negative interference, \colorbox{green1}{Blue} denotes positive synergy, with darker shades indicating better effects.}


\label{tab:interference_bilingual}       
\end{table*}

\subsection{Large-Scale Results on EC-30}\label{sec:ec30-results}

Similar to what we observed in the small-scale setting, we find notable improvements when we scale up on the EC30 dataset. 
As shown in Table~\ref{tab:ec30}, we show consistent improvements across high-, medium-, and low-resource languages, with an average gain of +1.3 SacreBLEU over the baseline. 
LaSS, while effective in high-resource O2M pairs, presents limitations with negative impacts (-1.0 score) on low-resource languages, highlighting difficulties in sub-network extraction for low-resource languages. 
In contrast, our method achieves stable and consistent gains across all resource levels. 
The Adapter$_{LP}$, despite increasing parameters by 87\% compared to the baseline, falls short of our method in boosting performance. 
Additionally, we show that applying Neuron Specialization in either the encoder or decoder delivers similar gains, with both combined offering stronger performance.

\begin{table}[H]
  \centering
  \resizebox{0.75\columnwidth}{!}{%
    \begin{tabular}{l|rrr}
      \toprule
      \multicolumn{1}{l|}{Model} & \multicolumn{1}{r}{$ \triangle \theta $} & \multicolumn{1}{r}{$ \triangle T_{subnet} $} & \multicolumn{1}{r}{$ \triangle $ Memory} 
      \\
      \midrule
      Adapter$_{\textit{LP}}$  & $+$87\% & n/a & 1.42 GB \\
      LaSS     & 0\% & $+$33 hours & 9.84 GB \\
      Ours     & 0\% & $+$5 minutes & 3e-3 GB \\
      \bottomrule
    \end{tabular}%
  }
  \caption{Efficiency comparison on EC30 dataset regarding extra trainable parameters ($ \triangle \theta $: relative increase over the baseline), extra processing time for subnet extraction ($ \triangle T_{subnet} $), and extra memory ($ \triangle $ Memory).
  }
  \label{tab:efficiency_comparisons}
\end{table}

\paragraph{Efficiency Comparisons.} We compare the efficiency on three aspects (Table~\ref{tab:efficiency_comparisons}). 
For trainable parameter increase, introducing lightweight language pair adapters accumulates a significant +87\% parameter growth over the baseline. 
Next, compared to LaSS, which is fine-tuned to identify sub-networks and demands substantial time (33 hours with 4 Nvidia A6000 GPUs), our approach efficiently locates specialized neurons in just 5 minutes. Considering memory costs, essential for handling numerous languages in deployment environments, our method proves more economical, primarily requiring storage of 1-bit masks for the FFN neurons instead of extensive parameters.

\paragraph{Random Mask.} We also incorporate the experiments using random masks with Neuron Specialization Training, to validate whether our Specialized Neuron Identification process can capture useful task-specific modularity. We randomly sample 70\% neurons to be task-specific and then conduct the same Neuron Specialization Training step. Our results indicate that the random masks strategy sacrifices performance on low-resource tasks (average -0.9 score) to enhance the performance of high-resource O2M directions (+0.9 score). This indicates the effectiveness of our identification method in locating intrinsic task-specific neurons.


\paragraph{The role of threshold factor.}\label{sec:hyper_k}
We explore the impact of our sole hyper-parameter $k$ (neuron selection threshold factor) on performance. 
The results indicate that performance generally improves with an increase in $k$, up to a point of 95\% (around 25\% sparsity), beyond which the performance starts to drop. 
See Appendix~\ref{appendix:hyper_k} for more detailed results.

\subsection{The Impact of Reducing Interference}\label{sec:interference_solved}


In this section, we evaluate to what extent our Neuron Specialization method mitigates interference and enhances cross-lingual transfer. Similar to~\citet{wang2020gradient}, we train bilingual models that do not contain interference or transfers, and then compare results between bilingual models, the conventional multilingual baseline model (mT-big), and our neuron specialization (ours). We train Transformer-big and Transformer-based models for high- and low-resource tasks, see Appendix~\ref{appendix:Model_Training_Details}. 

In Table~\ref{tab:interference_bilingual}, we show that the conventional multilingual model (mT-big) facilitates clear positive transfer for low-resource languages versus bilingual setups, leading to +8.2 (O2M) and +13.3 (M2O) score gains but incurs negative interference for high-resource languages (-3.7 and -1.1 scores).

Our method reduces interference for high-resource settings, leading to +1.8 and +1.4 SacreBLEU gains over mT-big in O2M and M2O directions. Moreover, our Neuron Specialization enhances low-resource language performance with average gains of +1.6 (O2M) and +1.2 (M2O) SacreBLEU over the mT-big, demonstrating its ability to foster cross-lingual transfer. Despite improvements, our approach still trails behind bilingual models for most high-resource O2M directions, indicating that while interference is largely reduced, room for improvement still exists.

\section{Conclusions}



In this paper, we have identified and leveraged \textit{intrinsic task-specific modularity} within multilingual networks to mitigate interference. We showed that FFN neurons activate in a language-specific way, and they present structural overlaps that reflect language proximity, which progress across layers. We then introduced \textit{Neuron Specialization} to leverage these natural modularity signals to structure the network, enhancing task specificity and improving knowledge transfer. Our experimental results, spanning various resource levels, show that our method consistently outperforms strong baseline systems, with additional analyses demonstrating reduced interference and increased knowledge transfer.
Our work deepens the understanding of multilingual models by revealing their intrinsic modularity, offering insights into how multi-task models can be optimized without extensive modifications.


\section*{Limitations}

This study primarily focuses on Multilingual Machine Translation, a key method in multi-task learning, using it as our primary testbed. 
However, the exploration of multilingual capabilities can be extended beyond translation to include a broader range of Multilingual Natural Language Processing tasks.
These areas remain unexplored in our current research and are considered promising directions for future work.

Additionally, our analysis is limited to the feed-forward network (FFN) components within the Transformer architecture, which, although they constitute a significant portion of the model's parameters, represent only one facet of its complex structure.
Future investigations could yield valuable insights by assessing the modularity of other Transformer components, such as the attention mechanisms or layer normalization modules, to provide a more comprehensive understanding of the system's overall functionality.

Lastly, we conducted our identification methods of specialized neurons primarily on Feed-Forward Networks that use ReLU as the activation function.
This is because neurons after the ReLU naturally present two states: active (>0) and inactive (=0), which offers a clear view of their contributions to the network outputs, thus being inherently interpretable. 
Recent work on Large Language Models has also explored the binary activation states of FFN neurons, particularly focused on when neurons are activated, and their roles in aggregating information~\cite{voita2023neurons}.
We leave the exploration of FFN neurons using other activation functions such as the GELU~\cite{hendrycks2016gaussian}, to future work.

\section*{Broader Impact}
Recognizing the inherent risks of mistranslation in machine translation data, we have made efforts to prioritize the incorporation of high-quality data, such as two open-sourced Multilingual Machine Translation datasets: IWSLT and EC30. Additionally, issues of fairness emerge, meaning that the capacity to generate content may not be equitably distributed across different languages or demographic groups. This can lead to the perpetuation and amplification of existing societal prejudices, such as biases related to gender, embedded in the data.

\bibliography{anthology,custom}
\bibliographystyle{acl_natbib}

\appendix

\section{Appendix}\label{sec:appendix}

\begin{table*}[h!]
\centering
\setlength{\tabcolsep}{3pt}
\renewcommand{\arraystretch}{1.85}
\begin{adjustbox}{width=\linewidth}
\begin{tabular}{c|ccc|ccc|ccc|ccc|ccc}
\toprule
 & \multicolumn{3}{c|}{Germanic} & \multicolumn{3}{c|}{Romance} & \multicolumn{3}{c|}{Slavic} & \multicolumn{3}{c|}{Indo-Aryan} & \multicolumn{3}{c}{Afro-Asiatic} \\ 
 & ISO & Language & Script & ISO & Language & Script & ISO & Language & Script & ISO & Language & Script & ISO & Language & Script \\ \hline
\multirow{2}{*}{\begin{tabular}[c]{@{}c@{}}High\\ (5m)\end{tabular}} & de & German & Latin & fr & French & Latin & ru & Russian & Cyrillic & hi & Hindi & Devanagari & ar & Arabic & Arabic \\
 & nl & Dutch & Latin & es & Spanish & Latin & cs & Czech & Latin & bn & Bengali & Bengali & he & Hebrew & Hebrew \\ \hline
\multirow{2}{*}{\begin{tabular}[c]{@{}c@{}}Med\\ (1m)\end{tabular}} & sv & Swedish & Latin & it & Italian & Latin & pl & Polish & Latin & kn & Kannada & Devanagari & mt & Maltese & Latin \\
 & da & Danish & Latin & pt & Portuguese & Latin & bg & Bulgarian & Cyrillic & mr & Marathi & Devanagari & ha & Hausa$^{*}$ & Latin \\ \hline
\multirow{2}{*}{\begin{tabular}[c]{@{}c@{}}Low\\ (100k)\end{tabular}} & af & Afrikaans & Latin & ro & Romanian & Latin & uk & Ukrainian & Cyrillic & sd & Sindhi & Arabic & ti & Tigrinya & Ethiopic \\
 & lb & Luxembourgish & Latin & oc & Occitan & Latin & sr & Serbian & Latin & gu & Gujarati & Devanagari & am & Amharic & Ethiopic \\
 \bottomrule
\end{tabular}
\end{adjustbox}
\centering 
\caption{Details of EC30 Training Dataset. Numbers in the table represent the number of sentences, for example, 5m denotes exactly 5,000,000 number of sentences. The only exception is Hausa, where its size is 334k (334,000).}
\label{tab:ec30-spec}
\end{table*}

\begin{table*}[h]
  \centering
  \resizebox{0.8\linewidth}{!}{
  \begin{tabular}{c|ccccccccc}
    \toprule
    \multirow{2}{*}{Models} & \multirow{2}{*}{Dataset} & \multicolumn{1}{c}{Num.} & \multicolumn{1}{c}{Num.} & \multicolumn{1}{c}{Num.} & \multirow{2}{*}{dim} & \multirow{2}{*}{$d_{\mathit{ff}}$} & \multicolumn{1}{c}{max} & \multicolumn{1}{c}{update} & \multirow{2}{*}{dropout}\\
     &  & trainable params & Layer & Attn Head &  &  & tokens & freq & \\
    \midrule
    mT-shallow  & IWSLT & 47M & 3 & 8 & 512 & 1,024  & 2,560 & 4 & 0.1 \\
    mT-small    & IWSLT & 76M & 6 & 8 & 512 & 1,024  & 2,560 & 4  & 0.1 \\
    \midrule
    bilingual-low  & EC30 & 52M  & 6 & 2 & 512 & 1,024 & 2,560  & 1 & 0.3 \\
    bilingual-high & EC30 & 439M & 6 & 16 & 1,024 & 4096 & 2,560  & 10 & 0.1 \\
    mT-big      & EC30   & 439M & 6 & 16 & 1,024 & 4,096 & 7,680  & 21 & 0.1 \\
    LaSS      & EC30   & 439M & 6 & 16 & 1,024 & 4,096 & 7,680  & 21 & 0.1 \\
    Neuron Specialization  & EC30 & 439M & 6 & 16 & 1,024 & 4,096 & 7,680  & 21 & 0.1 \\
    \bottomrule
  \end{tabular}}
  \caption{Configuration and hyper-parameter settings for all models in this paper. Num. Layer and Attn Head denote the number of layers and attention heads, respectively. dim represents the dimension of the Transformer model, $d_{\mathit{ff}}$ means the dimension of the feed-forward layer. bilingual-low and -high represent the bilingual models for low and high-resource languages.}
  \label{tab:model_spec}
\end{table*}

\subsection{Dataset details}\label{appendix:Dataset_Details}

\paragraph{IWSLT} We collect and pre-processes the IWSLT-14 dataset following~\citet{lin2021learning}. We refer readers to~\citet{lin2021learning} for more details.

\paragraph{EC30} We utilize the EC30, a subset of the EC40 dataset~\cite{tan2023towards} (with 10 extremely low-resource languages removed in our experiments) as our main dataset for most experiments and analyses. We list the Languages with their ISO and scripts in Table~\ref{tab:ec30-spec}, along with their number of sentences. In general, EC30 is an English-centric Multilingual Machine Translation dataset containing 61 million sentences covering 30 languages (excluding English). It collected data from 5 representative language families with multiple writing scripts. In addition, EC30 is well balanced at each resource level, for example, for all high-resource languages, the number of training sentences is 5 million. Note that the EC30 is already pre-processed and tokenized (with Moses tokenizer), thus we directly use it for our study.

\subsection{Model and Training Details}\label{appendix:Model_Training_Details}

We list the configurations and hyper-parameter settings of all systems for the main training setting (EC30) in Table~\ref{tab:model_spec}. As for global training settings, we adopt the pre-norm and share the decoder input output embedding for all systems. We use cross entropy with label smoothing to avoid overfitting (smoothing factor=0.1) and set early stopping to 20 for all systems. Similar to~\citet{fan2021beyond}, we prepend language tags to the source and target sentences to indicate the translation directions for all multilingual translation systems.

\paragraph{Bilingual models.} For bilingual models of low-resource languages, we adopt the suggested hyper-parameter settings from~\citet{araabi2020optimizing}, such as $d_{\mathit{ff}}=512$, number of attention head as 2, and dropout as 0.3. Furthermore, We train separate dictionaries for low-resource bilingual models to avoid potential overfitting instead of using the large 128k shared multilingual dictionary.

For bilingual models of high-resource languages, we adopt the 128k shared multilingual dictionary and train models with the Transformer-big architecture as the multilingual baseline (mT-big). The detailed configurations can be found in Table~\ref{tab:model_spec}.

\paragraph{Language Pair Adapters.} We implement Language Pair Adapters~\cite{bapna2019simple} by ourselves based on Fairseq. The Language Pair Adapter is learned depending on each pair, e.g., we learn two modules for en-de, namely en on the Encoder side and the de on the Decoder side. Note that, except for the unified pre-trained model, language pair adapters do not share any parameters with each other, preventing potential knowledge transfers. We set its bottleneck dimension as 128 for all experiments of IWSLT and EC30.

\begin{itemize}
    \item \textbf{IWSLT.} For the IWSLT dataset that contains 8 languages with 16 language pairs/translation directions, the size mT-small base model is 76M. Language Pair Adapters insert 3.2M additional trainable parameters for one language pair, thus resulting in 51.2M added parameters for all language pairs, leading to 67\% relative parameter increase over the baseline model.

    \item \textbf{EC30.} For the EC30 dataset that contains 30 languages with 60 language pairs/translation directions, the size mT-big base model is 439M. Language Pair Adapters insert 6.4M extra trainable parameters for one language pair, thus resulting in 384M added parameters for all language pairs, leading to 87\% relative parameter increase over the baseline model.
    \end{itemize}

\paragraph{Language Family Adapters. }The Language Family Adapter~\cite{chronopoulou2023language} is learned depending on each language family, e.g., for all 6 Germanic languages in the EC30, we learn two modules for en-Germanic, namely the en adapter on the Encoder side and the Germanic adapter on the Decoder side. We set its bottleneck dimension as 512 for all experiments for the EC30.

\begin{itemize}
   
    \item \textbf{EC30.} For the EC30 dataset that contains 30 languages with 60 language pairs/translation directions, the size mT-big base model is 439M. Language Family Adapters insert 25.3M additional trainable parameters for one family (on EN-X directions), thus resulting in 303.6M added parameters for all families on both EN-X and X-En directions, leading to 69\% relative parameter increase over the baseline model.
    \end{itemize}

\paragraph{LaSS.} When reproducing LaSS~\cite{lin2021learning}, we adopt the code from their official Github page\footnote{https://github.com/NLP-Playground/LaSS} with the same hyper-parameter setting as they suggested in their paper. For the IWSLT dataset, we finetune the mT-small for each translation direction with dropout=0.3, we then identify the language-specific parameters for attention and feed-forward modules (the setting with the strongest improvements in their paper) with a pruning rate of 70\%. We continue to train the sparse networks while keeping the same setting as the pre-training phase as they suggested. Note that we observed different results as they reported in the paper, even though we used the same code, hyper-parameter settings, and corresponding Python environment and package version. We also found that~\citet{he2023gradient} reproduced LaSS results in their paper, which shows similar improvements (around +0.6 BLUE gains) over the baseline of our reproductions. As for an improved method over LaSS proposed by~\citet{he2023gradient}, we do not reproduce their method since no open-source code has been released.

\subsection{Pseudocode of Neuron Specialization}\label{appendix:Pseudocodes}

We provide the pseudocode of our proposed method, \textit{Neuron Specialization}. We present the process of Specialized Neuron Identification in Algorithm. \ref{alg:NS_identification} and Neuron Specialization Training in Algorithm. \ref{alg:NS_training}.

\subsection{Result Details using ChrF++ and COMET}\label{appendix:Results}

For our main experiments in the EC30, we further provide the ChrF++~\cite{popovic2017chrf++} and COMET~\cite{rei2020comet} scores as extra results, as shown in Table~\ref{tab:ec30_chrf} and Table~\ref{tab:ec30_comet}, respectively. Similar to what we observed in Section~\ref{sec:ec30-results}, our Neuron Specialization presents consistent performance improvements over the baseline model while outperforming other methods such as LaSS and Adapters.

\begin{table*}[b]
\centering
\def\arraystretch{1.0}%
\resizebox{\linewidth}{!}{%
\begin{tabular}{c|c|ccccccccc|ccc}
\toprule
\multirow{2}{*}{Methods}& 
\multirow{2}{*}{$\Delta \theta$} & 
\multicolumn{3}{c}{High (5M)} & \multicolumn{3}{c}{Med (1M)} & \multicolumn{3}{c}{Low (100K)} & \multicolumn{3}{|c}{All (61M)} \\

\cmidrule(lr){3-5} \cmidrule(lr){6-8} \cmidrule(lr){9-11} \cmidrule(lr){12-14}
      & \multicolumn{1}{c|}{}
      & \multicolumn{1}{c}{O2M} & \multicolumn{1}{c}{M2O} & \multicolumn{1}{c}{Avg} 
      & \multicolumn{1}{c}{O2M} & \multicolumn{1}{c}{M2O} & \multicolumn{1}{c}{Avg} 
      & \multicolumn{1}{c}{O2M} & \multicolumn{1}{c}{M2O} & \multicolumn{1}{c}{Avg} 
      & \multicolumn{1}{|c}{O2M} & \multicolumn{1}{c}{M2O} & \multicolumn{1}{c}{Avg} \\
 \midrule

mT-big & - & 52.4 & 57.6 & 55.0 & 53.9 & 56.6 & 55.3 & 42.5 & 50.0 & 46.3 & 49.6 & 54.7 & 52.2 \\


Adapter$_{\textit{LP}}$ & +87\% & +1.3 & +0.2 & +0.8 & +1.1 & +0.1 & +0.6 & +0.3 & +0.3 & +0.3 & +0.9 & +0.2 & +0.5 \\

Adapter$_{\textit{Fam}}$ & +70\% & +0.6 & +0.2 & +0.4 & +0.7 & +0.3 & +0.5 & +1.1 & +0.4 & +0.8 & +0.8 & +0.3 & +0.5 \\

LaSS & 0\% & \textbf{+1.7} & +0.8 & +1.2 & \textbf{+1.3} & +0.3 & +0.8 & -0.3 & -1.5 & -0.9 & +0.9 & -0.2 & +0.5 \\

Random & 0\% & +0.7 & -0.4 & +0.2 & +0.4 & -0.5 & -0.1 & -0.5 & -1.2 & -0.9 & +0.2 & -0.7 & -0.3 \\

\midrule

Ours-Enc & 0\% & +1.0 & +0.9 & +1.0 & +0.7 & +0.9 & +0.8 & +0.6 & +0.9 & +0.8 & +0.8 & +0.9 & +0.8 \\

Ours-Dec & 0\% & +0.9 & +0.9 & +0.9 & +0.6 & +1.0 & +0.8 & +0.5 & +1.2 & +0.9 & +0.7 & +1.0 & +0.9 \\

Ours & 0\% & +1.3 & \textbf{+1.1} & \textbf{+1.2} & +1.1 & \textbf{+0.9} & \textbf{+1.0} & \textbf{+1.2} & \textbf{+0.8} & \textbf{+1.0} & \textbf{+1.2} & \textbf{+0.9} & \textbf{+1.1} \\

\bottomrule 
\end{tabular}%
}
\caption{Average \textbf{ChrF++} improvements on the EC30 dataset over the baseline (mT-big), categorized by High, Medium, and Low-resource translation directions. 'Ours-Enc' and 'Ours-Dec' indicate neuron specialization applied solely to the Encoder and Decoder, respectively, while 'Ours' signifies the method applied to both components. The best results are highlighted in \textbf{bold}.}
\label{tab:ec30_chrf}
\end{table*}

\begin{table*}[h!]
\centering
\def\arraystretch{1.0}%
\resizebox{\linewidth}{!}{%
\begin{tabular}{c|c|ccccccccc|ccc}
\toprule
\multirow{2}{*}{Methods}& 
\multirow{2}{*}{$\Delta \theta$} & 
\multicolumn{3}{c}{High (5M)} & \multicolumn{3}{c}{Med (1M)} & \multicolumn{3}{c}{Low (100K)} & \multicolumn{3}{|c}{All (61M)} \\

\cmidrule(lr){3-5} \cmidrule(lr){6-8} \cmidrule(lr){9-11} \cmidrule(lr){12-14}
      & \multicolumn{1}{c|}{}
      & \multicolumn{1}{c}{O2M} & \multicolumn{1}{c}{M2O} & \multicolumn{1}{c}{Avg} 
      & \multicolumn{1}{c}{O2M} & \multicolumn{1}{c}{M2O} & \multicolumn{1}{c}{Avg} 
      & \multicolumn{1}{c}{O2M} & \multicolumn{1}{c}{M2O} & \multicolumn{1}{c}{Avg} 
      & \multicolumn{1}{|c}{O2M} & \multicolumn{1}{c}{M2O} & \multicolumn{1}{c}{Avg} \\
 \midrule

mT-big & - & 83.4 & 83.9 & 83.65 & 81.1 & 80.1 & 80.6 & 73.8 & 73.4 & 73.6 & 79.1 & 79.1 & 79.1 \\

Adapter$_{\textit{LP}}$ & +87\% & +0.9 & +0.2 & +0.5 & +0.6 & +0.2 & +0.4 & 0 & +0.1 & 0 & +0.5 & +0.2 & +0.4 \\

Adapter$_{\textit{Fam}}$ & +70\% & +0.4 & +0.1 & +0.3 & +0.4 & +0.2 & +0.3 & +0.7 & +0.3 & +0.5 & +0.5 & +0.2 & +0.4 \\

LaSS & 0\% & \textbf{+1.5} & +0.8 & \textbf{+1.2} & \textbf{+0.9} & +0.6 & \textbf{+0.8} & -0.2 & -1.0 & -0.6 & +0.7 & +0.1 & +0.4 \\

Random & 0\% & +0.2 & -0.1 & +0.1 & -0.1 & -0.2 & -0.2 & -0.8 & -0.9 & -0.9 & -0.2 & -0.4 & -0.3 \\

\midrule

Ours-Enc & 0\% & +1.0 & +0.8 & +0.9 & +0.5 & +0.9 & +0.7 & +0.3 & \textbf{+0.9} & +0.6 & +0.6 & +0.8 & +0.7 \\

Ours-Dec & 0\% & +0.9 & +0.8 & +0.9 & +0.5 & \textbf{+1.0} & \textbf{+0.8} & +0.3 & \textbf{+0.9} & +0.6 & +0.6 & \textbf{+1.0} & +0.8 \\

Ours & 0\% & +1.4 & \textbf{+1.0} & \textbf{+1.2} & \textbf{+0.9} & +0.7 & \textbf{+0.8} & \textbf{+0.8} & +0.7 & \textbf{+0.8} & \textbf{+1.0} & +0.8 & \textbf{+0.9} \\

\bottomrule 
\end{tabular}%
}
\caption{Average \textbf{COMET} improvements on the EC30 dataset over the baseline (mT-big), categorized by High, Medium, and Low-resource translation directions. 'Ours-Enc' and 'Ours-Dec' indicate neuron specialization applied solely to the Encoder and Decoder, respectively, while 'Ours' signifies the method applied to both components. The best results are highlighted in \textbf{bold}.}
\label{tab:ec30_comet}
\end{table*}

\begin{figure}[h]
    \centering
    \includegraphics[width=\linewidth]{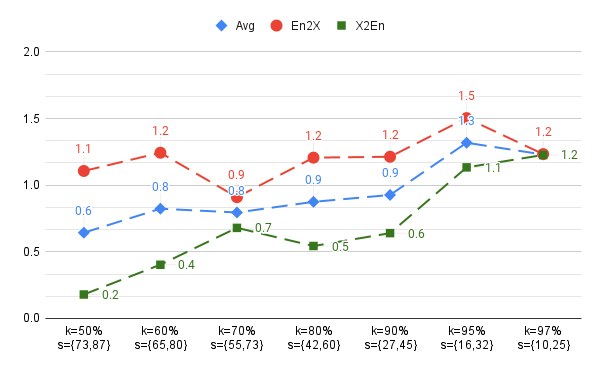}
    \caption{Improvements of Neuron Specialization method over the mT-large baseline on EC30. The x-axis indicates the factor $k$ and the dynamic sparsity of the fc1 layer, with displayed values ranging from minimum to maximum sparsity achieved. The y-axis indicates the SacreBLEU improvements over the mT-large model.} 
    \label{fig:k_sparsity_performance}
\end{figure}

\subsection{Sparsity versus Performance}\label{appendix:hyper_k}

For the Neuron Specialization, we dynamically select specialized neurons via a cumulative activation threshold $k$ in Equation~\ref{equation:threshold}, which is the only hyper-parameter of our method. Here, we discuss the impact of $k$ on the final performance and its relationship to the sparsity. As mentioned in Section~\ref{sec:Neuron_Analysis:Identification}, a smaller factor $k$ results in more sparse specialized neuron selection, which makes the fc1 weight more sparse as well in the Neuron Specialization Training process. In Figure~\ref{fig:k_sparsity_performance}, we show that increase $k$ leads to higher improvements in general, and the optimal performance is about when $k$=95\%. Such observation follows the intuition since when $k$ is too low, model capacity will be largely reduced.

\begin{figure}[h!]
    \centering
    \includegraphics[width=\linewidth]{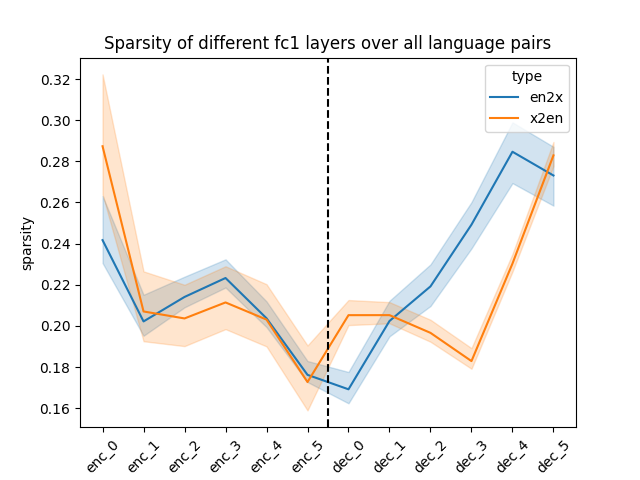}
    \caption{Sparsity progression of Neuron Specialization when $k=95$ on the EC30. We observe that the sparsity becomes smaller in the Encoder and then goes up in the Decoder. Note that this figure is based on the natural signals extracted from the untouched pre-trained model, and will be leveraged later in the process of Neuron Specialization Training.  This intrinsic pattern naturally follows our intuition that specialized neurons progress from language specific to agnostic the in Encoder, and vice versa in the Decoder.} 
    \label{fig:k95_sparsity}
\end{figure}

Furthermore, in Figure~\ref{fig:k95_sparsity}, we show that the sparsity of the network presents an intuitive structure: the sparsity decreases in the Encoder and increases in the Decoder. This implies the natural signal within the pre-trained multilingual model that neurons progress from language-specific to language-agnostic in the Encoder, and vice versa in the Decoder. Such observation is natural because it is reflected by the untouched network, similar to what we observed in the Progression of Neuron overlaps in Section~\ref{sec:Neuron_Structural_Analyasis:neuron_overlap_progression}.

\subsection{Visualization Details}\label{appendix:Visualizations}

\begin{figure*}[h!]
    \centering
    \includegraphics[width=0.8\linewidth]{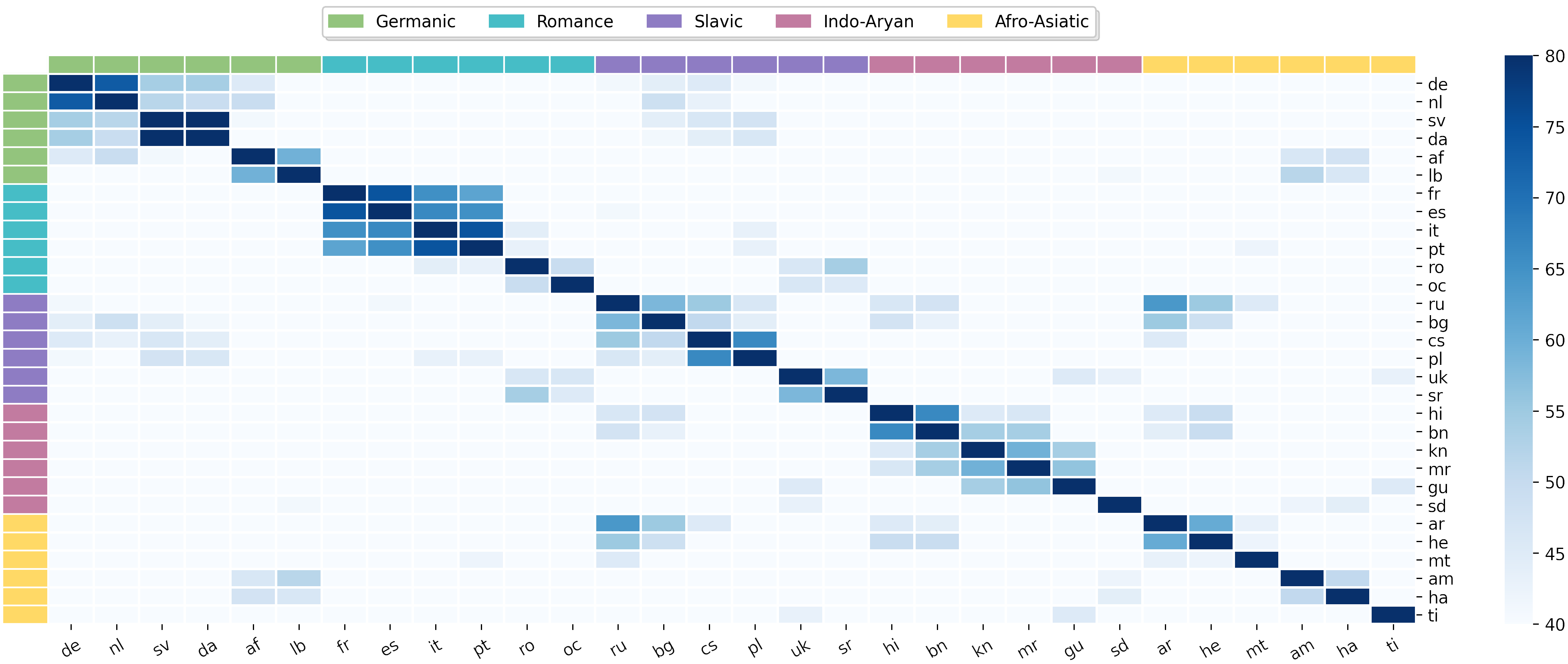}
    \caption{
    Pairwise Intersection over Union (IoU) scores for specialized neurons extracted from the \textbf{first encoder} FFN layer across all X-En language pairs to measure the degree of overlap between language pairs. Darker cells indicate stronger overlap, with the color threshold set from 40 to 80 to improve visibility.}
    \label{fig:iou_overall_enc_first}
\end{figure*}

\begin{figure*}[h!]
    \centering
    \includegraphics[width=0.8\linewidth]{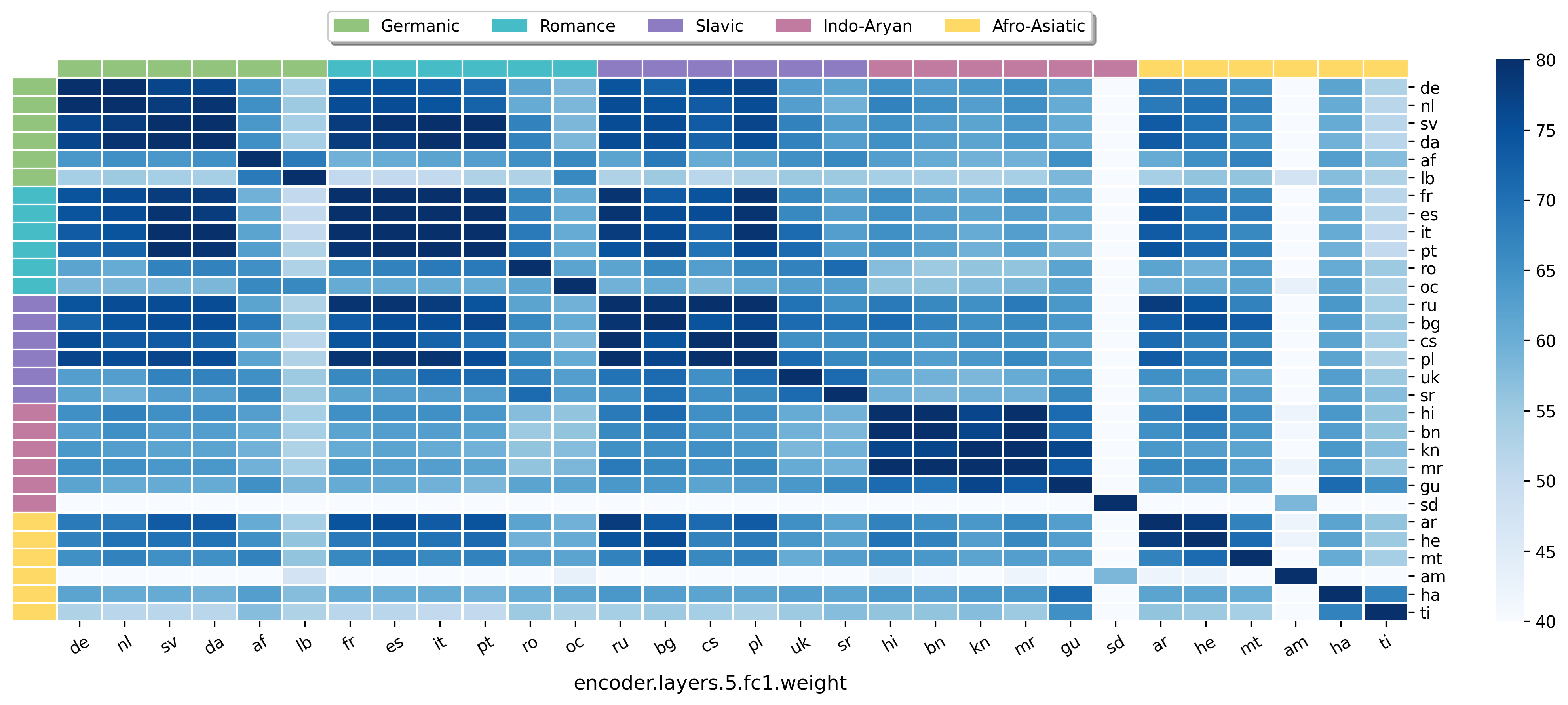}
    \caption{Pairwise Intersection over Union (IoU) scores for specialized neurons extracted from the \textbf{last encoder} FFN layer across all One-to-Many language pairs to measure the degree of overlap between language pairs. Darker cells indicate stronger overlap, with the color threshold set from 40 to 80 to improve visibility.}
    \label{fig:iou_overall_enc_last}
\end{figure*}

\begin{figure*}[h!]
    \centering
    \includegraphics[width=0.8\linewidth]{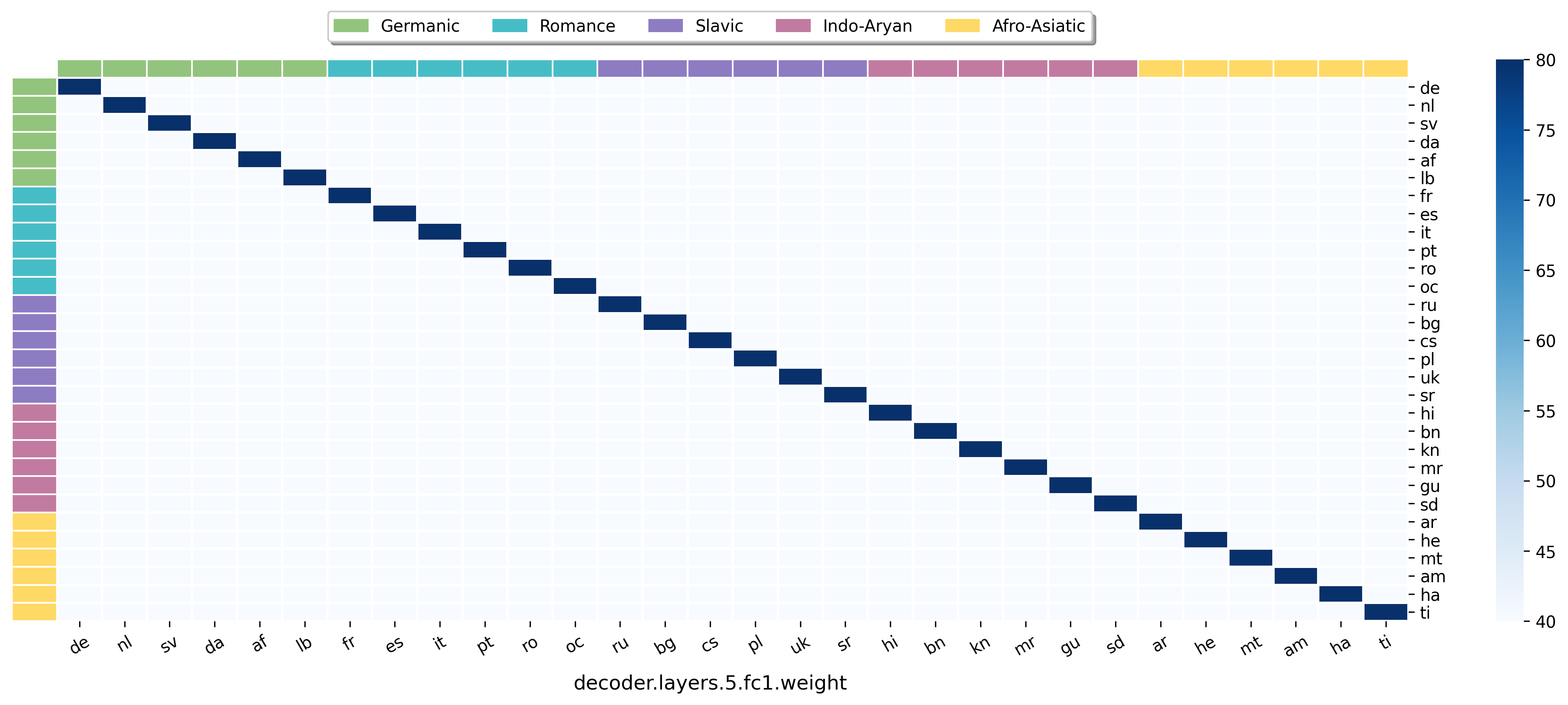}
    \caption{Pairwise Intersection over Union (IoU) scores for specialized neurons extracted from the \textbf{last decoder} FFN layer across all X-En language pairs to measure the degree of overlap between language pairs. Darker cells indicate stronger overlap, with the color threshold set from 40 to 80 to improve visibility.}
    \label{fig:iou_overall_dec_last}
\end{figure*}

We provide the additional Pairwise Intersection over Union (IoU) scores for specialized neurons in the first Encoder layer (Figure~\ref{fig:iou_overall_enc_first}), last Encoder layer (Figure~\ref{fig:iou_overall_enc_last}), and last Decoder layer (Figure~\ref{fig:iou_overall_dec_last}). The figures show that the Neurons gradually changed from language-specific to language-agnostic in the Encoder, and vice versa in the Decoder.

\begin{algorithm*}
\caption{Specialized Neuron Identification}
\label{alg:NS_identification}
\begin{algorithmic}[1]
\State \textbf{Input:} A pre-trained multi-task model $\theta$ with dimensions $d$ and $\mathit{d_{ff}}$; a validation dataset $D$ with $T$ tasks, where $D = \{D_1, ..., D_T\}$; and an accumulation threshold factor $k \in [0\%, 100\%]$ as the only hyper-parameter.
\State \textbf{Output:} A set of selected specialized neurons $S_k^t$ for each task $t$.
\For{task $t$ in $T$}
    \State \colorbox{gray!30}{Step 1: Activation Recording}
    \State Initialize activation vector $A_t = \mathbf{0} \in \mathbb{R}^{d_{\mathit{ff}}}$
    \For{sample $x_{i}$ in $D_{t}$}
        \State Record activation state $a^{t}_{i} \in \mathbb{R}^{d_{\mathit{ff}}}$
        \State $A_t = A_t + a^{t}_{i}$ \Comment{Accumulate activation states}
    \EndFor
    \State $a^{t} = \frac{A_t}{|D_{t}|}$ \Comment{Compute average activation state for task $t$}

    \State \colorbox{gray!30}{Step 2: Neuron Selection}
    \State Initialize selected neurons set $S_k^t = \emptyset$
    \While{selection condition not met} \Comment{Refer to Eq.~\ref{equation:threshold} for condition}
        \State Select neurons based on $a^t$ and add them to $S_k^t$
    \EndWhile
\EndFor
\end{algorithmic} 
\end{algorithm*}

\begin{algorithm*}
\caption{Neuron Specialization Training}
\label{alg:NS_training}
\begin{algorithmic}[1]
\State \textbf{Input:} A pre-trained multi-task model $\theta$ with dimensions $d$ and $\mathit{d_{ff}}$. Corpora data $C$ with $T$ tasks that contain both training and validation data. A set of selected specialized neurons $S_k^t$ for each task $t$.
\State \textbf{Output:} A new specialized network $\theta^{new}$. Note that only the fc1 weight matrix will be trained task-specifically, the other parameters are shared across tasks. In addition, $\theta^{new}$ does not contain more trainable parameters than $\theta$ due to the sparse network feature.
\State Derive boolean mask \mbox{$m^{t} \in \{0, 1\}^{d_{\mathit{ff}}}$} from $S_k^t$ for each layer
\While{$\theta^{new}$ not converge}
    \For{task $t$ in $T$}
        \State $W^{T}_{1} = m^{t} \cdot W^{\theta}_{1}$ \Comment{We perform this for all layers, refer to EQ.~\ref{equation:FFN_SN_transformer}}
        \State Train $\theta^{new}$ using $C^{t}$ \Comment{All parameters will be updated, yet fc1 layers are task specific}
    \EndFor
\EndWhile
\end{algorithmic} 
\end{algorithm*}

\end{document}